\documentclass[two column, journal]{IEEEtran}

\ifCLASSOPTIONcompsoc
  \usepackage[nocompress]{cite}
\else
  \usepackage{cite}
\fi

\ifCLASSINFOpdf
\else
\fi

\usepackage{amsmath}
\usepackage{amsfonts}
\usepackage{amssymb}
\usepackage{amsthm}
\usepackage{graphicx}
\newcommand{\bioheadshot}[1]{%
  \IfFileExists{#1}{\includegraphics[width=1in,height=1.25in,clip,keepaspectratio]{#1}}%
  {\fbox{\rule{0pt}{1.18in}\rule{0.88in}{0pt}}}}
\usepackage[caption=false,font=footnotesize]{subfig}
\usepackage{tabularx}
\usepackage{bm}
\usepackage{nicefrac}
\usepackage{microtype}
\usepackage[table]{xcolor}
\usepackage{booktabs,tabulary,multirow,overpic,calc}
\usepackage{colortbl}
\usepackage{enumitem}
\usepackage{makecell}
\usepackage{array}
\usepackage{wrapfig}
\usepackage{url}
\usepackage{float}
\usepackage[pagebackref=true,breaklinks=true,colorlinks,bookmarks=false,linkcolor=red,citecolor=cyan]{hyperref}


\definecolor{color1}{rgb}{0.95,0.95,0.95}
\definecolor{color2}{rgb}{0.858, 0.188, 0.478}
\definecolor{color3}{rgb}{0.95,0.95,0.95}
\definecolor{rouse}{rgb}{0.981,0.961,0.941}
\definecolor{light-yellow}{rgb}{1,1,0.93}
\definecolor{light-green}{rgb}{0.95,1,0.95}

\begin{document}

\title{Flow-Matching-Guided Deep Unfolding\\ for Hyperspectral Image Reconstruction}

\author{Yi~Ai,
        Yuanhao~Cai,
        Yulun~Zhang*,~%\IEEEmembership{Member,~IEEE,}
        Xin~Yuan,~%\IEEEmembership{Senior~Member,~IEEE,}
        and~Xiaokang~Yang,~\IEEEmembership{Fellow,~IEEE}%
\thanks{*Corresponding author: Yulun Zhang.}%
\thanks{Yi Ai, Yulun Zhang, and Xiaokang Yang are with School of Computer Science, Shanghai Jiao Tong University, China
(E-mail: ark\_ike@sjtu.edu.cn; yulun100@gmail.com; xkyang@sjtu.edu.cn).}%
\thanks{Yuanhao Cai is with Department of Computer Science, Johns Hopkins University, USA
(E-mail: caiyuanhao1998@gmail.com).}%
\thanks{Xin Yuan is with School of Engineering, Westlake University, China
(E-mail: xyuan@westlake.edu.cn).}%
% \thanks{X. Yang is with the School of Computer Science, Shanghai JiaoTong University, Shanghai 200240, China (e-mail: xkyang@sjtu.edu.cn).}
}

% The paper headers
\markboth{IEEE Transactions on Image Processing}%
{Yi Ai \MakeLowercase{\textit{et al.}}: Flow-Matching-Guided Deep Unfolding for Hyperspectral Image Reconstruction}

\IEEEtitleabstractindextext{%
\begin{abstract}
Reconstructing a three-dimensional hyperspectral cube from a two-dimensional compressed measurement is a severely ill-posed inverse problem. Existing deep unfolding networks (DUNs) retain fidelity to the imaging model, but regression-trained denoisers can suppress spatial detail and smooth spectral structure under strong modulation. This paper proposes \emph{FMU}, a deep unfolding framework that couples a measurement-conditioned flow-matching prior with a sensing-model-guided measurement update. A two-phase scheme first learns a compact clean-HSI latent target and then trains a conditional velocity field to generate this target from Gaussian noise. A mean-velocity regularizer additionally penalizes the first-moment error of the learned field. On the KAIST 10-scene benchmark under the optical-filter setting, FMU obtains 42.13\,dB PSNR and 0.9900 SSIM, outperforming LADE-DUN by 1.16\,dB in PSNR under the same training data, sensing mask, and evaluation protocol. Under the same optical-filter operator, FMU is further evaluated on held-out KAIST and ICVL scenes without fine-tuning. We also report quantitative simulated-CASSI results and qualitative reconstructions of real CASSI measurements.
\end{abstract}

}

\maketitle

\IEEEdisplaynontitleabstractindextext

\IEEEpeerreviewmaketitle

\ifCLASSOPTIONcompsoc
\IEEEraisesectionheading{\section{Introduction}\label{sec:introduction}}
\else
\section{Introduction}
\label{sec:introduction}
\fi

\IEEEPARstart{H}{yperspectral} imaging (HSI) resolves a scene across tens of narrow spectral bands and, in doing so, underpins tasks that RGB acquisition cannot serve---material identification, precision agriculture, biomedical diagnostics, and remote sensing~\cite{li2019deep, li2020prior, uzkent2017aerial, van2010tracking, rao2022siamese, goetz1985imaging, lu2020rafnet, stuart2019hyperspectral, rajabi2024hyperspectral, khan2022systematic, lu2014medical, ul2021review}. Because dense spectral sampling is expensive, snapshot compressive systems such as coded aperture snapshot spectral imaging (CASSI)~\cite{gehm2007single} and optical-filter-based imagers~\cite{yako2023video, xiong2022dynamic} multiplex the spatial and spectral dimensions onto a single two-dimensional sensor. Recovering the underlying $W\times H\times N_\lambda$ cube from such a measurement is a severely ill-posed inverse problem whose solvability depends almost entirely on the strength of the reconstruction prior.

\begin{figure}[!t]
\centering
\includegraphics[width=0.9\linewidth]{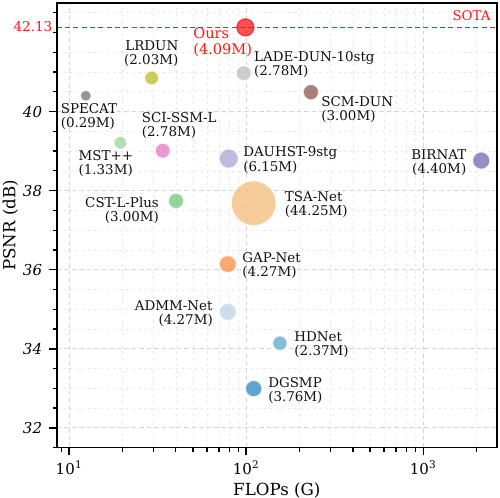}
\caption{Accuracy--complexity comparison on the optical-filter KAIST 10-scene benchmark at $256\times256$ resolution. FMU attains the highest average PSNR among the compared methods; marker area indicates parameter count.}
\label{fig:cmp}
\end{figure}

Unlike scanning spectrometers, snapshot systems capture spatial and spectral content in a single exposure and are therefore attractive for dynamic scenes and compact instruments. This advantage comes at the price of aggressive multiplexing. In CASSI, a coded aperture and a dispersive element superimpose wavelength-dependent, spatially shifted observations. In an optical-filter camera, spectral channels are mixed at the same detector coordinates according to calibrated filter responses. Although the two systems instantiate different sensing operators, both retain far fewer observations than unknown cube elements. A reconstruction algorithm must obey the sensing model and use an expressive prior to resolve measurement ambiguities.

Early methods imposed sparsity, total variation, or low-rank structure and solved the resulting optimization problem iteratively~\cite{wagadarikar2008single,kittle2010multiframe,wang2016adaptive,liu2018rank}. Their explicit objectives provide physical interpretability, but handcrafted priors have limited capacity for diverse spatial textures and spectral signatures. End-to-end CNN, recurrent, and Transformer models later learned a direct measurement-to-cube mapping~\cite{meng2020end,hu2022hdnet,cheng2022recurrent,cai2022mask,cai2022coarse}. These networks substantially improve accuracy, yet the sensing operator is often used only during measurement preprocessing or loss construction. Severe spectral mixing can therefore leave multiple plausible reconstructions.

Deep unfolding networks bridge model-based optimization and data-driven representation learning. They translate a finite number of iterations into network stages, using a projection or gradient step to enforce measurement consistency and a learned denoiser to model the image prior~\cite{ma2019deep,meng2023deep,wang2022snapshot,cai2022degradation}. Representative designs improve the proximal module through spectral attention, degradation awareness, low-rank modeling, or latent generative information~\cite{dong2023residual,li2023pixel,meng2023deep,wu2024latent}. Nevertheless, the prior used by most denoisers is learned with point-estimation losses. Such supervision favors an average solution when the inverse mapping is highly ambiguous, which may attenuate fine textures and smooth narrow spectral variations. Generative priors must balance distribution modeling with computational cost.

The difficulty is especially pronounced in spectral imaging because spatial appearance alone does not determine the underlying spectrum. Materials with similar brightness may exhibit distinct spectral responses, while narrow absorption features can be easily removed as noise by a regression denoiser. Conversely, hallucinating visually plausible texture without respecting the measurement can corrupt material signatures. A useful generative prior must therefore be conditioned on the captured measurement and coupled tightly with data consistency. It should guide reconstruction toward the clean-HSI manifold while retaining the measured evidence.

How the prior is incorporated is equally important. Invoking a generative sampler independently at every unfolding stage multiplies its cost by the number of stages and makes the quality--efficiency tradeoff prohibitive. Directly generating the full hyperspectral cube also duplicates the role of the reconstruction network and increases the dimensionality of generative modeling. We instead seek a compact conditional representation that can be generated once, shared across stages, and decoded jointly with measurement-consistent features. The generative model thus captures clean spectral--spatial statistics, while unfolding reconstructs the full cube.

Diffusion models have demonstrated the value of generative priors for inverse problems~\cite{ho2020denoising,song2020score}. Their iterative reverse process, however, introduces a quality--latency tradeoff: reducing the number of denoising steps improves speed but may weaken the generated prior. Flow matching offers a different construction~\cite{lipman2022flow,liu2022flow}. It directly learns a continuous velocity field that transports a simple source distribution toward the target distribution and generates samples by solving an ODE. An explicit velocity target and flexible numerical integration suit the generation of a compact, stage-shared prior.

Motivated by these observations, we propose a flow-matching-guided deep unfolding network, termed FMU. A learned measurement-update module uses the sensing operator and measurement residual to refine the stage input. A conditional flow model generates a compact clean-HSI latent from the measurement back-projection, and a multi-scale denoiser injects this latent into every unfolding stage. The prior is sampled once and shared across stages, avoiding repeated generative sampling within the reconstruction loop. Training is organized into two phases: the first learns an informative latent target from clean hyperspectral data, while the second learns the measurement-conditioned velocity field. A mean-velocity constraint complements pointwise flow regression by penalizing the channel-wise aggregate velocity error.

The framework is evaluated on both optical-filter-based and CASSI systems. Beyond the ten-scene KAIST benchmark, we report spectral fidelity, complexity, held-out KAIST and ICVL generalization, simulated CASSI reconstruction, and qualitative real-measurement results. Our contributions are:
\begin{itemize}[leftmargin=*]
    \item We develop a physics-driven unfolding framework guided by a measurement-conditioned latent flow. The generated clean-HSI-derived prior is shared across reconstruction stages, while each stage uses a sensing-model-based residual update before prior-guided denoising.
    \item We introduce a mean-velocity regularizer that penalizes the channel-wise first-moment error of the learned vector field. It complements samplewise prior matching with negligible computation and improves reconstruction in our ablations.
    \item Under the same training data, sensing mask, and evaluation protocol, FMU achieves 42.13\,dB PSNR and 0.9900 SSIM on the optical-filter KAIST benchmark, exceeding LADE-DUN by 1.16\,dB in PSNR. Additional experiments report SAM and ERGAS, evaluation on held-out KAIST and ICVL scenes under the same sensing operator, quantitative simulated-CASSI results, and qualitative real-CASSI reconstructions.
\end{itemize}

\begin{figure*}[!t]
    \centering
    \subfloat[CASSI system\label{fig:sys_cassi}]{%
        \includegraphics[width=0.48\textwidth]{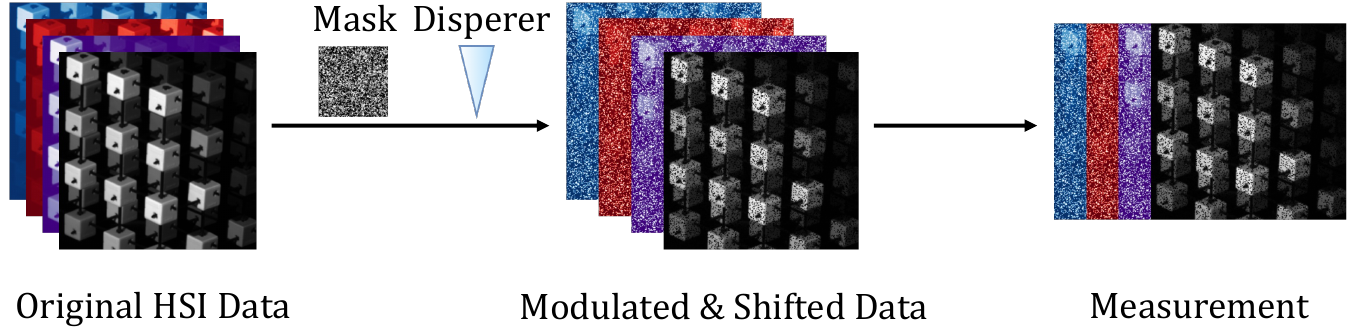}}
    \hfill
    \subfloat[Optical-filter-based HSI system\label{fig:sys_filter}]{%
        \begin{overpic}[width=0.48\textwidth]{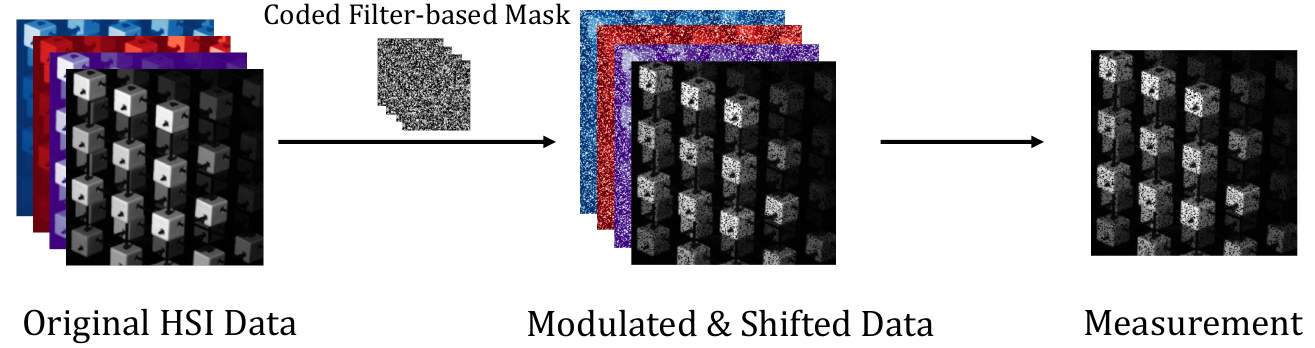}
          \put(32,0){\colorbox{white}{\scriptsize Spectrally Modulated Data}}
        \end{overpic}}
    \caption{Two representative snapshot HSI acquisition modalities. \emph{Left}: a classical CASSI system separates spatial and spectral encoding through a coded aperture followed by a dispersive prism, producing a spatially shifted mask that varies with wavelength. \emph{Right}: a filter-based system encodes both dimensions with a single broadband optical filter, yielding a simpler, chip-integrable optical path but a more entangled inverse problem for reconstruction algorithms to solve.}
    \label{fig:overall}
\end{figure*}

\section{Related Work}

\subsection{Snapshot Hyperspectral Imaging Systems}
HSI acquisition has evolved from pushbroom and whiskbroom scanning toward snapshot instruments~\cite{ortega2019hyperspectral,uto2016development,baek2017compact}. Scanning systems acquire spatial or spectral samples sequentially and can provide high spectral resolution, but their acquisition time makes them sensitive to scene motion. Snapshot systems encode a cube in one or a few exposures, shifting the acquisition burden to computational reconstruction.

CASSI is a representative snapshot architecture~\cite{gehm2007single}. In its single-disperser configuration~\cite{wagadarikar2008single}, a coded aperture modulates the scene before a disperser shifts each wavelength by a different amount on the detector. The resulting measurement contains spatial--spectral shearing and overlap among adjacent bands. This forward model has motivated many reconstruction algorithms and provides a standard simulation and real-data testbed for spectral compressive imaging.

Optical-filter-based cameras encode wavelength-dependent transmittance at the sensor plane using, for example, Fabry--P\'{e}rot filter arrays~\cite{yako2023video}. Other compact designs use birefringent spectral demultiplexing~\cite{marois2023birefringent} or combine spectral filtering with diffuser-induced spatial mixing~\cite{monakhova2020spectral}. These are distinct encoding mechanisms within the development of miniaturized spectrometers~\cite{yang2021miniaturization}. In our optical-filter setting, calibrated transmittances modulate the bands without wavelength-dependent spatial displacement, mixing spectral contributions at each spatial coordinate. We formulate FMU for a general linear sensing operator and evaluate both this aligned optical-filter setting and dispersive CASSI with their corresponding forward and adjoint operators.

\subsection{Hyperspectral Image Reconstruction}
\textbf{Model-based methods.} Classical reconstruction combines an explicit data-fidelity term with a handcrafted regularizer. Sparse coding represents hyperspectral signals in a learned or analytic basis~\cite{wagadarikar2008single,kittle2010multiframe,wang2016adaptive}, while low-rank models exploit correlations across spectral bands~\cite{liu2018rank}. Total variation and related smoothness priors suppress noise and artifacts. These methods are interpretable and can operate without large training sets, but their limited priors cannot fully describe complicated textures and material spectra. Iterative optimization also entails careful parameter selection and substantial computational cost.

\textbf{End-to-end networks.} Data-driven methods learn the inverse mapping from simulated measurement--cube pairs. CNN approaches such as TSA-Net and HDNet build spatial--spectral feature hierarchies~\cite{meng2020end,hu2022hdnet}, while BIRNAT models spectral dependencies recurrently~\cite{cheng2022recurrent}. Transformer-based MST, MST++, CST-L-Plus, and SPECAT use attention to capture long-range spatial and inter-band correlations~\cite{cai2022mask,cai2022mstpp,cai2022coarse,yao2024specat}, and SCI-SSM-L adopts spatial--spectral state-space modeling~\cite{shi2026scissm}. Their strong representation capacity produces high reconstruction accuracy, but a purely feed-forward mapping does not explicitly correct the estimate against the physical measurement at every depth. Mask or noise changes can expose this distinction.

\textbf{Deep unfolding.} Deep unfolding maps the iterations of an optimization algorithm to a finite network~\cite{ma2019deep,meng2023deep,wang2022snapshot,cai2022degradation}. Each stage typically contains an analytic data-consistency update followed by a learned proximal or denoising module. Existing designs strengthen this learned component through degradation estimation, spectral attention, residual degradation learning, generative latent information, Mamba-based proximal models, and low-rank physical models~\cite{dong2023residual,li2023pixel,meng2023deep,wu2024latent,feng2025scmdun,huang2026lrdun}. These developments show that accuracy depends not only on the projection formula, but also on how effectively each stage obtains and uses prior information. FMU follows this model-driven principle and learns a conditional latent distribution that supplies a common high-quality prior to all stages.

\subsection{Diffusion and Flow-Matching Generative Priors}
Diffusion models~\cite{ho2020denoising,song2020score} learn to reverse a prescribed corruption process and have become powerful priors for inverse imaging. For hyperspectral reconstruction, a diffusion-generated latent can complement the measurement-driven features of an unfolding network~\cite{wu2024latent}. The reverse trajectory nevertheless requires repeated network evaluations, and its cost depends strongly on the sampler and number of denoising steps. This motivates generative formulations that retain distribution modeling while offering a more direct transport path.

Flow matching (FM)~\cite{lipman2022flow} instead learns a continuous-time vector field that transports a simple source density to the data distribution. Under a linear-interpolation coupling, training reduces to supervised regression of the target velocity. Rectified Flow studies straighter transport trajectories~\cite{liu2022flow}, while optimal-transport variants improve the source--target coupling~\cite{kornilov2024optimal}. Once trained, the vector field defines a deterministic ODE and can be integrated with fixed- or adaptive-step solvers such as RK45~\cite{dormand1980family}. FM therefore separates learned transport dynamics from the choice of numerical accuracy at inference.

Most generative image models synthesize pixels directly. For compressive HSI reconstruction, generating a compact latent is more economical and permits the physics-driven network to retain responsibility for the final cube. Conditioning the velocity field on the back-projected measurement further aligns the generated latent with the observed scene rather than drawing an unconditional sample. FMU builds on these properties: it transports Gaussian noise toward a clean-HSI latent target, injects the resulting conditional prior at multiple feature scales, and regularizes the aggregate velocity in addition to the conventional pointwise objective.

\begin{figure*}[!t]
  \centering
  \includegraphics[width=0.95\linewidth]{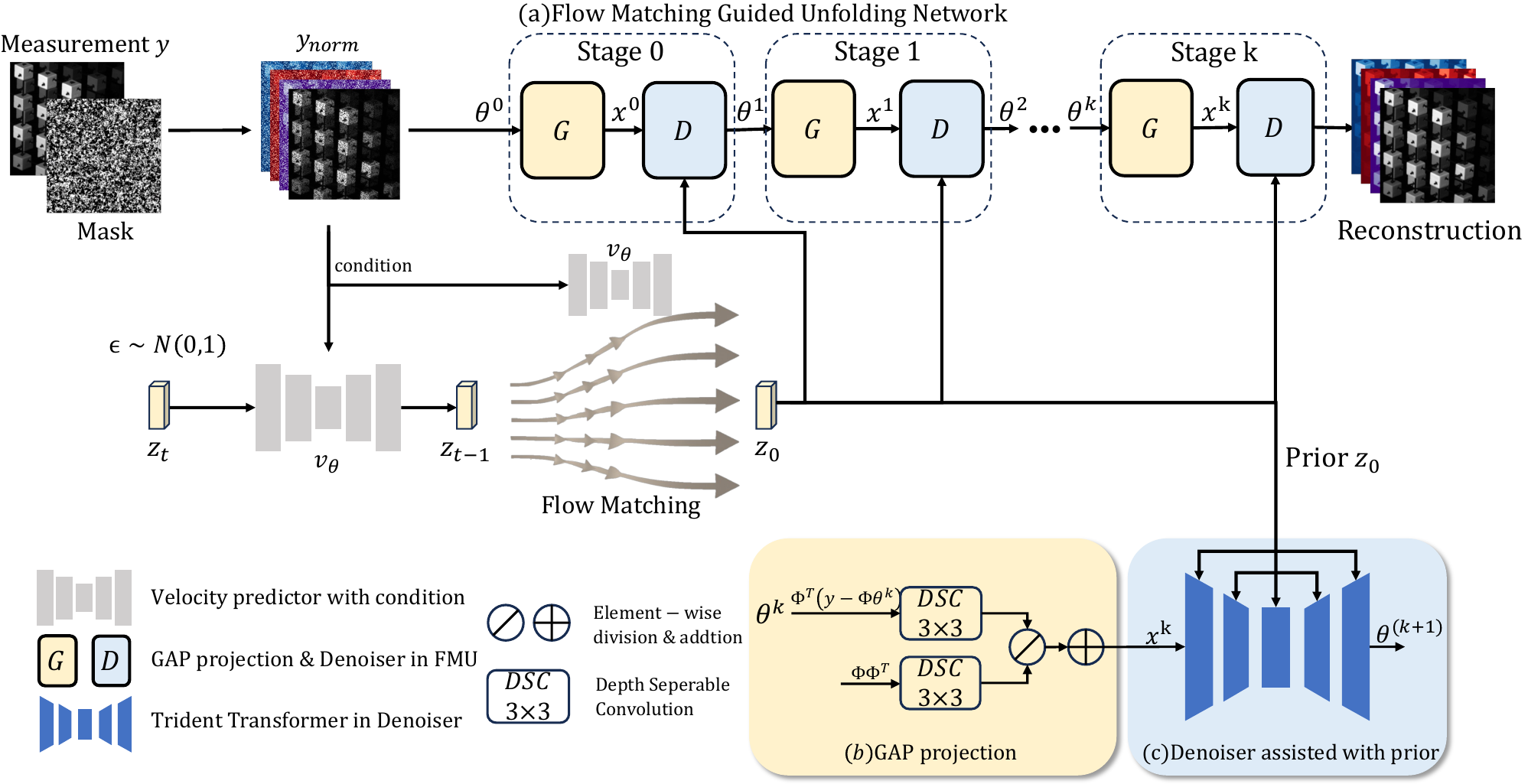}
  \caption{Overall pipeline of the proposed FMU. The compressed measurement $\bm y$ enters an $N$-stage unfolding network. Each stage combines a GAP-motivated measurement update with a U-shaped denoiser composed of Trident Transformer blocks. The update uses the forward and adjoint sensing operators to correct the measurement residual. The flow-matching prior $\bm z_{FM}$---generated once from $\bm y$ via a flow-matching ODE---is injected into every denoising stage through cross-attention, so the same clean-HSI-derived prior guides refinement at every scale. The prior is sampled once outside the unfolding loop.}
  \label{fig:flow}
\end{figure*}

\section{Method}
\subsection{Modeling of the HSI System}
The CASSI architecture acquires hyperspectral information by encoding spectral slices and projecting them onto a 2D detector. Let the hyperspectral cube be
\(\mathbf{X} \in \mathbb{R}^{W \times H \times N_\lambda}\),
where \(W\), \(H\), and \(N_\lambda\) denote spatial width, height, and the number of spectral channels. A mask \(\mathbf{m} \in \mathbb{R}^{W \times H}\) first modulates each spectral slice, after which the disperser shifts the modulated slice by $(\ell-1)d$ pixels along the horizontal detector axis. With zero padding outside the valid support, the measurement is
\begin{equation}
\mathbf{Y}_{\mathrm{CASSI}} = \sum_{\ell=1}^{N_\lambda}
\mathcal S_{(\ell-1)d}\!\left(\mathbf{X}_{:,:,\ell}\odot\mathbf m\right)
+\mathbf{N}_{\mathrm{CASSI}},
\end{equation}
where \(\odot\) denotes the Hadamard product and $\mathcal S_s$ shifts its argument by $s$ pixels. The resulting detector has size $W\times(H+(N_\lambda-1)d)$. Equivalently, one may precompute a three-dimensional shifted mask, but the wavelength-dependent displacement is applied only once.

An optical-filter system modulates each spectral band with its calibrated transmittance mask, giving
\begin{equation}
\mathbf{Y} = \sum_{n_\lambda=1}^{N_\lambda}\mathbf{X}(:,:,n_\lambda) \odot \mathbf{M}(:,:,n_\lambda)+\mathbf{N},
\end{equation}
where \(\mathbf{Y}\in \mathbb{R}^{W \times H}\), \(\mathbf M\in \mathbb{R}^{W \times H\times N_\lambda}\) is the calibrated spectral-transmittance mask, and \(\mathbf N\) denotes noise. Without wavelength-dependent displacement, this operator multiplexes all bands at the same spatial locations on the detector.

For both systems, vectorizing the cube and measurement gives $\bm{x}=\mathrm{vec}(\mathbf X)\in\mathbb R^{WHN_\lambda}$ and $\bm y=\mathrm{vec}(\mathbf Y)\in\mathbb R^{n}$, where $n=WH$ for the optical-filter operator and $n=W(H+(N_\lambda-1)d)$ for CASSI. The forward model becomes
\begin{equation}
\label{eq:convex}
\bm{y} = \bm\Phi \bm{x} + \bm{\eta},
\end{equation}
where $\bm\eta\in\mathbb R^n$ is measurement noise and $\bm\Phi\in\mathbb R^{n\times WHN_\lambda}$ denotes the modality-specific sensing operator. For the optical-filter system, $\bm\Phi=[\mathbf D_1,\ldots,\mathbf D_{N_\lambda}]$ with $\mathbf D_\ell=\mathrm{Diag}(\mathrm{vec}(\mathbf M(:,:,\ell)))$; for CASSI, each block additionally contains the corresponding shift operator $\mathcal S_{(\ell-1)d}$.
Thus, hyperspectral image reconstruction can be viewed as solving the ill-posed linear inverse problem of reconstructing \(\mathbf{x}\) from \(\mathbf{y}\).

In both systems, $\bm\Phi$ can be applied through mask multiplication and band summation, with an additional spatial shift for CASSI; the full sensing matrix need not be stored. Its adjoint distributes the detector residual back to the spectral bands using the same mask and, for CASSI, reverse shifting and cropping. These operations provide a common interface for the measurement update while retaining the distinct acquisition geometries of the two systems.

\subsection{Proposed Model}
We propose a flow-matching-guided unfolding network (FMU) (Fig.~\ref{fig:flow}). The measurement is processed by an $N$-stage deep unfolding network (DUN), comprising a measurement-update module and a denoiser in each stage. The U-shaped Transformer denoiser uses Trident Transformer blocks~\cite{wu2024latent} to combine updated features with the flow-matching prior.

\subsubsection{Flow-Matching-Guided Unfolding Framework}
We adopt the generalized alternating projection (GAP) framework~\cite{liao2014generalized} to motivate the reconstruction stages. For Eq.~(\ref{eq:convex}), consider the regularized objective for the unknown hyperspectral cube
\begin{equation}
    \hat{\bm{x}}=\underset{\bm x}{\arg\min}\ \frac 12\|\bm y-\bm\Phi\bm x\|^2+\tau R(\bm x).
\end{equation}
The objective enforces consistency with the compressed measurement, while $\tau R(\bm x)$ encodes an image prior. To expose the standard GAP update, we use the idealized noise-free constraint associated with Eq.~(\ref{eq:convex}); for noisy measurements, the resulting projection serves as a data-consistency step rather than an assertion that $\bm\eta=\bm 0$. An auxiliary variable $\bm\theta$ separates measurement consistency from prior-driven refinement:
\begin{equation}
    (\hat{\bm{x}},\hat{\bm{\theta}})=\underset{\bm {x,\theta}}{\arg\min}\ \frac 12\|\bm x-\bm\theta\|^2+\tau R(\bm \theta),\ \ \text{s.t.}\ \bm y=\bm{\Phi}\bm x.
\end{equation}
The update scheme alternates between two operations. Euclidean projection enforces measurement consistency and data fidelity by projecting the intermediate solution onto the feasible set. The prior-guided denoising step then leverages learned priors to refine the reconstruction:
\begin{equation}
\begin{aligned}
\bm x^{(k+1)}&=\bm \theta^{(k)} + \bm\Phi^\dagger \big(\bm y - \bm\Phi \bm \theta^{(k)}\big),\\
\bm \theta^{(k+1)} &= \mathcal{F}_{k+1}\!\left(\bm x^{(k+1)}; \bm z_{FM} \right).
\end{aligned}
\label{eq:gap_denoise}
\end{equation}
Here, $\mathcal{F}_{k+1}$ denotes a flow-matching-guided denoiser in the $(k+1)$-th stage, and $\bm z_{FM}$ is the flow-matching prior learned from clean HSIs (see Sec.~\ref{sec:FM}). This prior provides strong generative regularization and enables the model to better capture spectral--spatial correlations, leading to improved recovery even with a limited number of unfolding stages.
For the diagonalized snapshot operator used in implementation, $\bm\Phi^\dagger\bm r$ is stabilized as $\bm\Phi^\top\bm r/(\bm\Phi\bm\Phi^\top+\epsilon)$ with element-wise division and a small $\epsilon$.
Equation~(\ref{eq:gap_denoise}) provides the optimization motivation for the stage-wise architecture. Following the GC-GAP design in LADE-DUN~\cite{wu2024latent}, the implemented network replaces the analytic projection with a learned gradient-correction (GC) update. Let $\mathbf P_{\Phi}\in\mathbb R^{H_d\times W_d\times N_\lambda}$ denote the operator-aligned sensing tensor that parameterizes $\bm\Phi$, and let $\mathbf Y\in\mathbb R^{H_d\times W_d}$ be the unvectorized measurement. The dimensions $(H_d,W_d)$ equal the detector dimensions of the corresponding sensing system. We use $\mathcal A_{\mathbf P}$ and $\mathcal A_{\mathbf P}^{\top}$ for the forward and adjoint operations induced by $\mathbf P$. At stage $k$, the module predicts a residual correction to the sensing tensor, refines the measurement residual, and supplies the corrected feature to the denoiser:
\begin{align}
\widetilde{\mathbf P}^{(k)}&=\mathbf P_{\Phi}
+\mathcal C_{\Phi}^{(k)}([\mathbf Y,\mathbf P_{\Phi}]),\\
\widetilde{\mathbf R}^{(k)}&=\mathcal C_{r}^{(k)}([\mathcal A_{\widetilde{\mathbf P}^{(k)}}(\bm\theta^{(k)})-\mathbf Y,
\mathcal A_{\widetilde{\mathbf P}^{(k)}}(\bm\theta^{(k)}),\mathbf Y]),\\
\bm g^{(k+1)}&=\mathcal C_{g}^{(k)}([\bm\theta^{(k)},
\mathcal A_{\widetilde{\mathbf P}^{(k)}}^{\top}(\widetilde{\mathbf R}^{(k)})]),\\
\bm\theta^{(k+1)}&=\mathcal F_{k+1}
\!\left(\bm g^{(k+1)};\bm z_{FM}\right).
\label{eq:gradient_correction}
\end{align}
Here, $[\cdot]$ denotes channel concatenation after the operands are arranged in channel-first tensor form, with $\mathbf Y$ providing one measurement channel. $\mathcal C_{\Phi}$ consists of two pointwise--depthwise--pointwise convolution blocks, $\mathcal C_r$ is a $1\times1$ convolution, and $\mathcal C_g$ contains two depthwise--pointwise convolution blocks. Equation~(\ref{eq:gap_denoise}) gives the optimization motivation, while Eq.~(\ref{eq:gradient_correction}) specifies the complete learned stage update used in the network.

\subsubsection{Training Strategy of FMU}
\label{sec:FM}
\begin{figure*}[!t]
  \centering
  \includegraphics[width=0.9\linewidth]{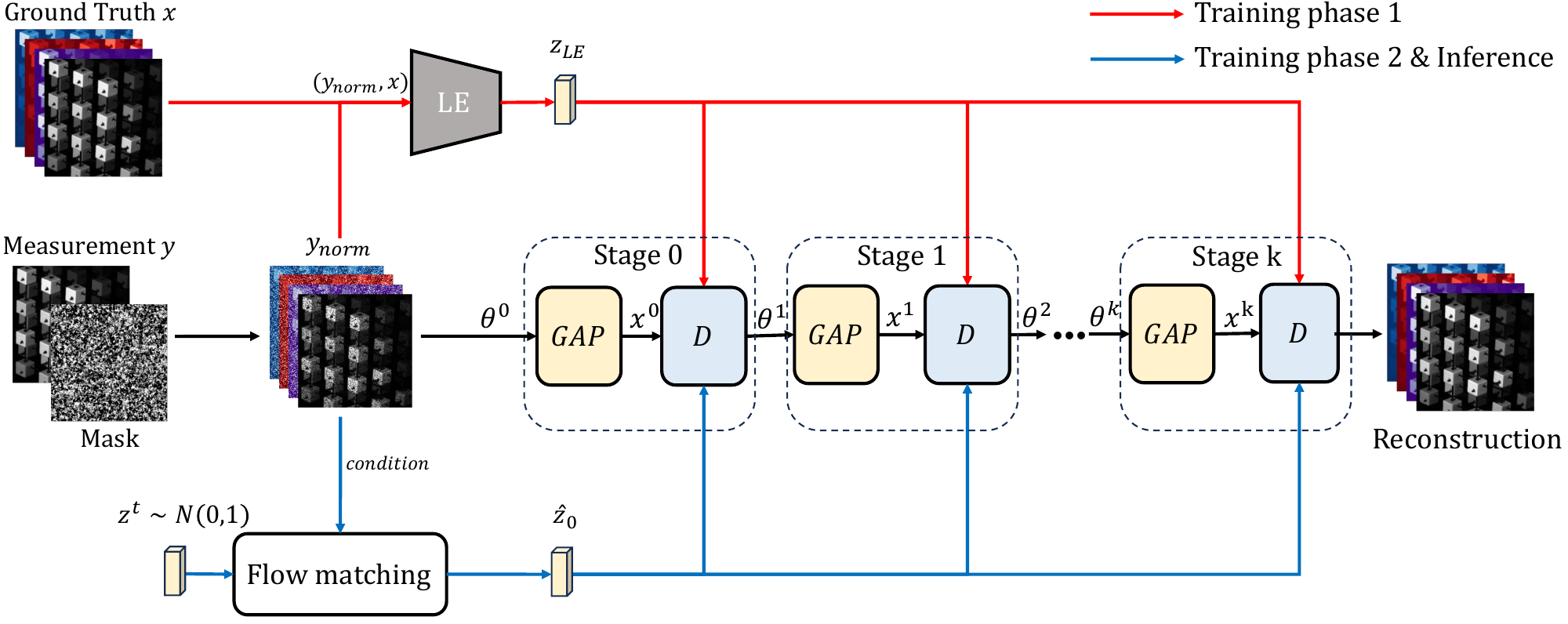}
  \caption{Two-phase training procedure. Phase~1 jointly learns the clean-HSI latent encoder and unfolding reconstruction network using the reconstruction loss. Phase~2 freezes this latent encoder, uses its output as the flow target, and jointly optimizes the conditional flow and the remaining reconstruction modules.}
  \label{fig:training}
\end{figure*}

The two-phase procedure in Fig.~\ref{fig:training} jointly trains the flow prior and unfolding denoisers.

\textbf{In the first phase}, a clean-HSI latent target is learned from the ground-truth cube and its measurement back-projection.
The latent encoder and the complete unfolding reconstruction network are optimized jointly. The encoder takes the ground-truth HSI $\bm x$ and the measurement back-projection as input.
Before encoding, we compute the normalized measurement back-projection
$\bm y_{norm}=\bm\Phi^\dagger\bm y\in\mathbb{R}^{W\times H\times N_\lambda}$.
The encoder input is
$I_E=\mathrm{concat}(\bm y_{norm},\bm x)\in\mathbb{R}^{W\times H\times 2N_\lambda}$,
and the latent prior is obtained as
$\bm z_{LE}=\mathrm{LE}(I_E)\in\mathbb{R}^{Q\times C}$,
where $Q$ and $C$ denote the token and channel dimensions, respectively.
The encoder first applies a pixel unshuffle to reduce spatial resolution while increasing channel dimensionality,
followed by MobileBlocks~\cite{howard2019searching} for efficient local feature learning,
and an MLP-Mixer module~\cite{tolstikhin2021mlp} for effective channel-wise and spatial mixing.
The detailed architecture of the latent encoder is provided in Section~\ref{app:architecture}.
The output $\bm z_{LE}$ is then used as the prior to guide denoising, leading to the reconstruction $\hat{\bm x}=\mathrm{FMU}(\bm y,\bm \Phi,\bm z_{LE})$.
We supervise the latent encoder and unfolding network with the reconstruction loss
\begin{equation}
\mathcal L_1=\mathcal L_{rec}=\rho(\hat{\bm x}-\bm x),
\end{equation}
where $\rho(\bm e)=\operatorname{mean}\sqrt{\bm e^2+\epsilon^2}$ is the Charbonnier penalty, evaluated elementwise and averaged with $\epsilon=10^{-3}$.

This joint objective trains the latent code through its contribution to reconstruction. Gradients from $\mathcal L_{rec}$ reach the encoder through the prior-injection modules, so the latent features are learned together with the denoisers that consume them. The ground-truth input supplies information absent from the compressed measurement, while the back-projection associates that information with the observed degradation. Freezing the encoder in Phase~2 preserves this target representation as the conditioning encoder, velocity predictor, and reconstruction modules are optimized to generate and use a prior from the compressed measurement alone.

\textbf{In the second phase}, the flow-matching model learns to generate the latent prior conditioned on \(\bm y_{norm}\).
We initialize the reconstruction network from Phase~1, freeze only the clean-HSI encoder \(\mathrm{LE}\), and use its output \(\bm z_{LE}\) as the flow target. The measurement-conditioning encoder, velocity predictor, GC modules, and unfolding denoisers remain trainable and are jointly refined by the Phase~2 objective.
Following the notation in Figs.~\ref{fig:flow} and~\ref{fig:training}, let
$\bm z_T\sim p_T=\mathcal N(0,I)$ denote the terminal Gaussian state and
$\bm z_0=\bm z_{LE}\sim p_0$ the clean-HSI latent target. Sampling follows the reverse-time direction $T\rightarrow0$, parameterized by $s=1-t/T\in[0,1]$, where $s=0$ corresponds to $\bm z_T$ and $s=1$ to $\bm z_0$. The measurement back-projection $\bm y_{norm}$ conditions the velocity predictor, whereas the ground-truth HSI is used only by the frozen latent encoder to construct $\bm z_0$.
The flow solves
\begin{equation}
    \frac{\mathrm d\bm z_s}{\mathrm ds}=v_\theta(\bm z_s,s,\bm y_{norm}),\qquad s\in[0,1].
\end{equation}
The linear interpolation is
$\bm z_s=(1-s)\bm z_T+s\bm z_0$, whose constant target velocity is $\bm z_0-\bm z_T$. We first write the conventional squared regression objective as
\begin{equation}
\label{eq:8}
    \mathcal L_{vel}=\mathbb E_{s,\bm z_T,\bm z_0}\left[
    \|\bm z_0-\bm z_T-v_\theta(\bm z_s,s,\bm y_{norm})\|^2_2\right],
\end{equation}
where $s$ is sampled uniformly. Rather than optimizing Eq.~(\ref{eq:8}) directly, our implementation uses its robust endpoint counterpart. It forms
$\tilde{\bm z}_0=\bm z_T+v_\theta(\bm z_s,s,\bm y_{norm})$ and applies a Charbonnier penalty between $\tilde{\bm z}_0$ and $\bm z_{LE}$. Since their difference equals
$v_\theta-(\bm z_0-\bm z_T)$, this endpoint objective replaces the squared velocity loss while preserving the same target. At inference, RK45 integrates $s$ from approximately $0$ to $1$ to produce $\hat{\bm z}_{FM}$ for all unfolding stages.

During flow training, $\bm z_s$ contains a sampled mixture of noise and the target latent, while the measurement condition remains fixed for that training pair. Sampling $s$ exposes the predictor to different noise--target proportions along the prescribed path. At inference, the target latent is unavailable, and the state evolves using the learned field conditioned on the observed measurement throughout the integration interval.

\subsubsection{Mean-Velocity Constraint}
\label{sec:mean_velocity}
The pointwise endpoint objective is estimated from sampled Gaussian--latent pairs, times, and latent tokens. Its channel-wise residual can fluctuate across these samples and tokens. We therefore introduce a coupling-preserving regularizer that leaves the sampled paths unchanged and directly penalizes this aggregate residual. Let the latent velocity tensor have batch, token, and channel axes $(b,q,c)$. The \emph{mean-velocity constraint} averages over the batch and token axes and retains a channel-wise first moment:
\begin{equation}
\begin{aligned}
\bar v_{\theta,c}&=\frac{1}{BQ}\sum_{b,q}[v_\theta]_{bqc},\\
\bar v_c^{\ast}&=\frac{1}{BQ}\sum_{b,q}[\bm z_0-\bm z_T]_{bqc},\\
\mathcal{L}_{mean}&=\frac{1}{C}\sum_{c=1}^{C}
\left(\bar v_{\theta,c}-\bar v_c^{\ast}\right)^2.
\end{aligned}
\end{equation}
Both moments are estimated from the same minibatch, so the regularizer adds only two reductions and a channel-wise squared difference. The endpoint prior loss fits individual source--target pairs, whereas $\mathcal L_{mean}$ explicitly penalizes their aggregate channel-wise velocity bias.

To interpret this term, let $e_{bqc}=[v_\theta-(\bm z_0-\bm z_T)]_{bqc}$. The regularizer is the squared channel-wise mean of $e_{bqc}$: residuals with a consistent sign within a channel accumulate, while opposite-signed residuals can cancel. The endpoint loss still penalizes these individual errors. Averaging includes the $Q$ latent tokens, so the aggregate constraint remains defined when training with a single cube per minibatch.

The second-phase objective combines reconstruction, robust endpoint matching, and the mean-velocity regularizer:
\begin{align}
\mathcal{L}_{2} &= \mathcal L_{rec}+\mathcal L_{prior}
 +\lambda_{mean}\mathcal{L}_{mean},\\
\mathcal L_{prior} &= \rho(\tilde{\bm z}_0-\bm z_{LE}).
\end{align}
where $\rho$ is the Charbonnier penalty used in our implementation and $\lambda_{mean}$ trades samplewise prior matching against channel-wise aggregate consistency. In Sec.~\ref{sec:exp} we show that a moderate $\lambda_{mean}$ (empirically $\lambda_{mean}=5$) delivers the best test PSNR, while $\lambda_{mean}=0$ and $\lambda_{mean}=100$ both underperform.

\subsubsection{Prior Sampling and Reconstruction}
At inference, the measurement back-projection supplies the conditioning features, and a Gaussian state initializes the latent trajectory. Integrating the conditional velocity field gives
\begin{equation}
\hat{\bm z}_{FM}=\bm z_T+\int_0^1
v_\theta(\bm z_s,s,\bm y_{norm})\,\mathrm ds.
\label{eq:prior_sampling}
\end{equation}
The integrated prior is reused in all $N$ updates of Eq.~(\ref{eq:gradient_correction}); the clean-HSI encoder is absent from this inference path. Here, $\tilde{\bm z}_0$ denotes the endpoint estimate used in the training loss, whereas $\hat{\bm z}_{FM}$ denotes the ODE output used for reconstruction. The number of velocity evaluations is determined by the RK45 solver tolerance, independently of the unfolding depth.

During this process, the latent code remains fixed while the image estimate and its measurement residual evolve across stages. Each denoiser combines the shared code with its current measurement-updated input. This separates the cost of conditional prior generation from the repeated refinement of the full-resolution hyperspectral cube.

\subsection{Network Architecture}
\label{app:architecture}
For a $256\times256$ input, the clean-HSI encoder concatenates the 28-channel back-projection and 28-channel ground truth. A $4\times4$ pixel-unshuffle, MobileBlocks~\cite{howard2019searching}, adaptive $4\times4$ pooling, and an MLP-Mixer~\cite{tolstikhin2021mlp} produce $\bm z_{LE}\in\mathbb R^{16\times256}$. The measurement-conditioning encoder has the same structure but takes only the 28-channel back-projection. For each latent token, the SimpleCNN velocity predictor concatenates the 256-D state, 256-D condition, and scalar time, applies two length-preserving 1-D convolutions with 512 channels, and outputs a 256-D velocity for the sampled state.

The flow therefore evolves 4,096 latent values, compared with 1,835,008 voxel values in a $256\times256\times28$ cube. This reduction confines iterative prior sampling to a compact representation. Full-resolution spatial--spectral reconstruction is performed by the unfolding denoisers, which receive both the measurement-updated features and the generated code.

The reconstruction module uses 10 unfolding stages. The first and last stages have separate parameters, whereas the eight intermediate stages share one GC module and one denoiser. The denoiser uses three feature scales of 28, 56, and 112 channels. Following~\cite{wu2024latent}, the generated $16\times256$ latent is reduced to 16, 4, and 1 tokens and injected at the corresponding scales through cross-prior interaction.

\begin{table*}[!t]
\caption{Optical-filter reconstruction results (PSNR/SSIM) on ten KAIST scenes and their average. Best and second-best values are shown in bold and underlined, respectively. Parameter counts correspond to the evaluated configurations.}
\label{table:hsi_scene10_avg}
\scriptsize
\setlength{\tabcolsep}{0.75mm}
\renewcommand\arraystretch{1.12}
\centering
\resizebox{\textwidth}{!}{
\begin{tabular}{ccccccccccccc|c}
\toprule
\rowcolor{color3}
Method & Venue & ~~Params~~ & ~~Scene1~~ & ~~Scene2~~ & ~~Scene3~~ & ~~Scene4~~ & ~~Scene5~~ & ~~Scene6~~ & ~~Scene7~~ & ~~Scene8~~ & ~~Scene9~~ & ~~Scene10~~ & ~~Avg~~ \\
\midrule
$\lambda$-Net~\cite{miao2019net} & ICCV'19 & 62.64M
& \makecell{32.62\\0.9247} & \makecell{31.43\\0.9003} & \makecell{30.49\\0.9109} & \makecell{32.34\\0.9185}
& \makecell{31.57\\0.9391} & \makecell{33.44\\0.9410} & \makecell{32.98\\0.9092} & \makecell{29.22\\0.9075}
& \makecell{31.33\\0.9218} & \makecell{30.90\\0.9178} & \makecell{31.63\\0.9191} \\
\midrule
DGSMP~\cite{huang2021deep} & CVPR'21 & 3.76M
& \makecell{34.09\\0.9495} & \makecell{30.83\\0.9188} & \makecell{30.41\\0.9182} & \makecell{38.05\\0.9654}
& \makecell{32.74\\0.9584} & \makecell{35.20\\0.9719} & \makecell{33.96\\0.9349} & \makecell{31.20\\0.9510}
& \makecell{31.39\\0.9371} & \makecell{31.99\\0.9666} & \makecell{32.99\\0.9472} \\
\midrule
HDNet~\cite{hu2022hdnet} & CVPR'22 & 2.37M
& \makecell{33.73\\0.9523} & \makecell{33.46\\0.9416} & \makecell{34.14\\0.9555} & \makecell{40.18\\0.9736}
& \makecell{33.15\\0.9649} & \makecell{35.70\\0.9713} & \makecell{34.71\\0.9445} & \makecell{31.13\\0.9463}
& \makecell{32.66\\0.9495} & \makecell{32.53\\0.9659} & \makecell{34.14\\0.9565} \\
\midrule
ADMM-Net~\cite{ma2019deep}& ICCV'19 & 4.27M
& \makecell{35.57\\0.9622} & \makecell{35.22\\0.9539} & \makecell{34.18\\0.9473} & \makecell{39.80\\0.9716}
& \makecell{34.67\\0.9642} & \makecell{36.11\\0.9675} & \makecell{34.46\\0.9353} & \makecell{31.82\\0.9475}
& \makecell{33.76\\0.9506} & \makecell{33.66\\0.9698} & \makecell{34.93\\0.9570} \\
\midrule
BiSRNet~\cite{cai2023binarized}& NeurIPS'23 & 0.036M
& \makecell{36.23\\0.9628} & \makecell{35.79\\0.9526} & \makecell{34.11\\0.9289} & \makecell{40.05\\0.9563}
& \makecell{35.55\\0.9632} & \makecell{37.34\\0.9653} & \makecell{35.27\\0.9381} & \makecell{34.44\\0.9512}
& \makecell{34.71\\0.9392} & \makecell{34.95\\0.9660} & \makecell{35.84\\0.9524} \\
\midrule
GAP-Net~\cite{meng2023deep}& IJCV'23 & 4.27M
& \makecell{36.35\\0.9674} & \makecell{37.16\\0.9615} & \makecell{35.85\\0.9568} & \makecell{41.17\\0.9799}
& \makecell{35.89\\0.9701} & \makecell{36.45\\0.9747} & \makecell{35.90\\0.9523} & \makecell{32.26\\0.9537}
& \makecell{35.98\\0.9609} & \makecell{34.39\\0.9754} & \makecell{36.14\\0.9653} \\
\midrule
TSA-Net~\cite{meng2020end}& ECCV'20 & 44.25M
& \makecell{37.87\\0.9745} & \makecell{39.51\\0.9809} & \makecell{36.08\\0.9641} & \makecell{43.72\\0.9852}
& \makecell{36.43\\0.9799} & \makecell{38.06\\0.9821} & \makecell{36.71\\0.9596} & \makecell{35.18\\0.9738}
& \makecell{37.30\\0.9762} & \makecell{35.89\\0.9830} & \makecell{37.68\\0.9759} \\
\midrule
CST-L-Plus~\cite{cai2022coarse}& ECCV'22 & 3.00M
& \makecell{37.42\\0.9736} & \makecell{38.99\\0.9801} & \makecell{37.88\\0.9697} & \makecell{42.41\\0.9800}
& \makecell{36.94\\0.9819} & \makecell{38.08\\0.9828} & \makecell{36.88\\0.9681} & \makecell{35.39\\0.9718}
& \makecell{36.18\\0.9720} & \makecell{37.19\\0.9820} & \makecell{37.74\\0.9762} \\
\midrule
MST++~\cite{cai2022mstpp}& CVPRW'22 & 1.33M
& \makecell{39.19\\0.9815} & \makecell{41.58\\0.9871} & \makecell{39.40\\0.9759} & \makecell{44.36\\0.9893}
& \makecell{38.63\\0.9865} & \makecell{39.15\\0.9870} & \makecell{38.27\\0.9727} & \makecell{35.49\\0.9756}
& \makecell{39.33\\0.9832} & \makecell{36.66\\0.9859} & \makecell{39.21\\0.9825} \\
\midrule
BIRNAT~\cite{cheng2022recurrent}& TPAMI'23 & 4.40M
& \makecell{38.72\\0.9787} & \makecell{40.83\\0.9839} & \makecell{39.25\\0.9757} & \makecell{43.46\\0.9854}
& \makecell{37.92\\0.9832} & \makecell{38.83\\0.9838} & \makecell{37.30\\0.9632} & \makecell{35.90\\0.9747}
& \makecell{38.88\\0.9799} & \makecell{36.50\\0.9840} & \makecell{38.76\\0.9792} \\
\midrule
DAUHST-9stg~\cite{cai2022degradation}& NeurIPS'22 & 6.15M
& \makecell{38.26\\0.9785} & \makecell{40.28\\0.9850} & \makecell{38.26\\0.9748} & \makecell{44.04\\0.9859}
& \makecell{37.87\\0.9869} & \makecell{39.23\\0.9852} & \makecell{38.44\\0.9750} & \makecell{35.66\\0.9746}
& \makecell{38.62\\0.9814} & \makecell{37.40\\0.9881} & \makecell{38.81\\0.9815} \\
\midrule
SPECAT~\cite{yao2024specat}& CVPR'24 & 0.29M
& \makecell{39.81\\0.9849}
& \makecell{42.18\\0.9892}
& \makecell{40.71\\0.9798}
& \makecell{44.57\\0.9878}
& \makecell{39.45\\0.9895}
& \makecell{40.31\\0.9902}
& \makecell{40.08\\0.9800}
& \makecell{37.60\\0.9816}
& \makecell{41.15\\0.9866}
& \makecell{38.14\\0.9889}
& \makecell{40.40\\0.9859} \\
\midrule
LADE-DUN-10stg~\cite{wu2024latent}& ECCV'24 & 2.78M
& \makecell{40.10\\0.9853}
& \makecell{43.27\\0.9919}
& \makecell{41.76\\0.9849}
& \makecell{\textbf{47.55}\\\underline{0.9949}}
& \makecell{39.76\\0.9906}
& \makecell{40.43\\0.9910}
& \makecell{39.94\\0.9809}
& \makecell{37.72\\\underline{0.9833}}
& \makecell{41.21\\0.9897}
& \makecell{37.94\\0.9895}
& \makecell{40.97\\0.9882} \\
\midrule
SCM-DUN-9stg~\cite{feng2025scmdun}& TCSVT'26 & 3.00M
& \makecell{40.17\\0.9840}
& \makecell{42.72\\0.9897}
& \makecell{42.56\\0.9841}
& \makecell{45.85\\0.9936}
& \makecell{39.59\\0.9891}
& \makecell{39.40\\0.9892}
& \makecell{39.39\\0.9787}
& \makecell{36.66\\0.9794}
& \makecell{40.92\\0.9881}
& \makecell{37.64\\0.9887}
& \makecell{40.49\\0.9865} \\
\midrule
SCI-SSM-L~\cite{shi2026scissm}& PR'26 & 2.78M
& \makecell{38.97\\0.9810}
& \makecell{40.95\\0.9868}
& \makecell{38.78\\0.9747}
& \makecell{44.62\\0.9887}
& \makecell{38.82\\0.9864}
& \makecell{38.68\\0.9861}
& \makecell{38.20\\0.9726}
& \makecell{35.38\\0.9738}
& \makecell{39.27\\0.9820}
& \makecell{36.40\\0.9851}
& \makecell{39.01\\0.9817} \\
\midrule
LRDUN-9stg~\cite{huang2026lrdun}& CVPR'26 & 2.03M
& \makecell{40.25\\0.9848}
& \makecell{43.43\\0.9921}
& \makecell{42.49\\0.9851}
& \makecell{46.69\\0.9933}
& \makecell{39.90\\0.9903}
& \makecell{39.85\\0.9903}
& \makecell{39.67\\0.9803}
& \makecell{37.17\\0.9823}
& \makecell{41.32\\0.9889}
& \makecell{37.71\\0.9884}
& \makecell{40.85\\0.9876} \\
\midrule
\rowcolor{light-green}
\textbf{FMU-S (Ours)} & -- & 2.78M
& \makecell{\underline{41.42}\\\underline{0.9877}}
& \makecell{\underline{44.28}\\\underline{0.9927}}
& \makecell{\underline{43.29}\\\underline{0.9860}}
& \makecell{46.75\\\underline{0.9949}}
& \makecell{\underline{40.77}\\\underline{0.9918}}
& \makecell{\underline{41.00}\\\underline{0.9917}}
& \makecell{\textbf{41.69}\\\underline{0.9850}}
& \makecell{\underline{37.83}\\0.9828}
& \makecell{\textbf{43.11}\\\underline{0.9906}}
& \makecell{\underline{39.21}\\\underline{0.9905}}
& \makecell{\underline{41.94}\\\underline{0.9894}} \\
\midrule
\rowcolor{light-green}
\textbf{FMU (Ours)}& -- & 4.09M
& \makecell{\textbf{41.84}\\\textbf{0.9883}}
& \makecell{\textbf{44.29}\\\textbf{0.9930}}
& \makecell{\textbf{43.54}\\\textbf{0.9869}}
& \makecell{\underline{47.14}\\\textbf{0.9955}}
& \makecell{\textbf{41.15}\\\textbf{0.9926}}
& \makecell{\textbf{41.15}\\\textbf{0.9922}}
& \makecell{\underline{41.65}\\\textbf{0.9851}}
& \makecell{\textbf{38.17}\\\textbf{0.9840}}
& \makecell{\underline{42.85}\\\textbf{0.9910}}
& \makecell{\textbf{39.49}\\\textbf{0.9913}}
& \makecell{\textbf{42.13}\\\textbf{0.9900}} \\
\bottomrule
\end{tabular}
}
\end{table*}

\section{Experiments}
\subsection{Experiment Settings}
\label{sec:exp}
Consistent with~\cite{cai2022mask}, the evaluation uses 28 spectral channels with wavelengths ranging from 450 to 650\,nm and covers both simulated and real measurements.

\textbf{Simulated Optical-Filter-Based HSI System Data.} We use CAVE~\cite{yasuma2010generalized} and KAIST~\cite{choi2017high}. CAVE contains 32 original $512\times512$ hyperspectral images. Following the standard SCI preprocessing, the training set comprises 205 indexed 28-band, $1024\times1024$ cubes (\emph{scene1}--\emph{scene205}). Evaluation uses the fixed ten $256\times256$ KAIST scenes distributed with the TSA-Net simulation benchmark (\emph{scene01}--\emph{scene10}). Following SPECAT~\cite{yao2024specat}, measurements are formed using the same calibrated $256\times256\times28$ Fabry--P\'{e}rot filter mask~\cite{yako2023video}.

\textbf{Simulated CASSI System Data.} We use the same CAVE/KAIST training and test splits as above, using the dispersive CASSI operator with a two-pixel spectral shift in place of the aligned optical-filter mask.

\textbf{Real CASSI System Data.} For real hyperspectral data, we adopt a dataset collected using the CASSI system, which was previously introduced in TSA-Net~\cite{meng2020end}. To align with real acquisition conditions, all training samples are generated as simulated CASSI measurements, with additional noise incorporated to approximate the sensor projection process.

\textbf{Metrics and Implementation Specifics.} We use peak signal-to-noise ratio (PSNR) and structural similarity index (SSIM) as the primary reconstruction metrics. The model is implemented in PyTorch~\cite{paszke2019pytorch} and trained on one NVIDIA A6000 GPU with 48\,GB memory. Both phases run for 300 epochs with Adam ($\beta_{1}=0.9$, $\beta_{2}=0.999$), an initial learning rate of $4\times 10^{-4}$, and a MultiStepLR schedule with milestones $\{50,100,150,200,250\}$ and decay factor $0.5$. Further settings and spectral metrics appear in Secs.~\ref{sec:implementation-details} and~\ref{subsec:metrics_fairness}.

\begin{figure*}[!t]
\centering
\includegraphics[width=\linewidth]{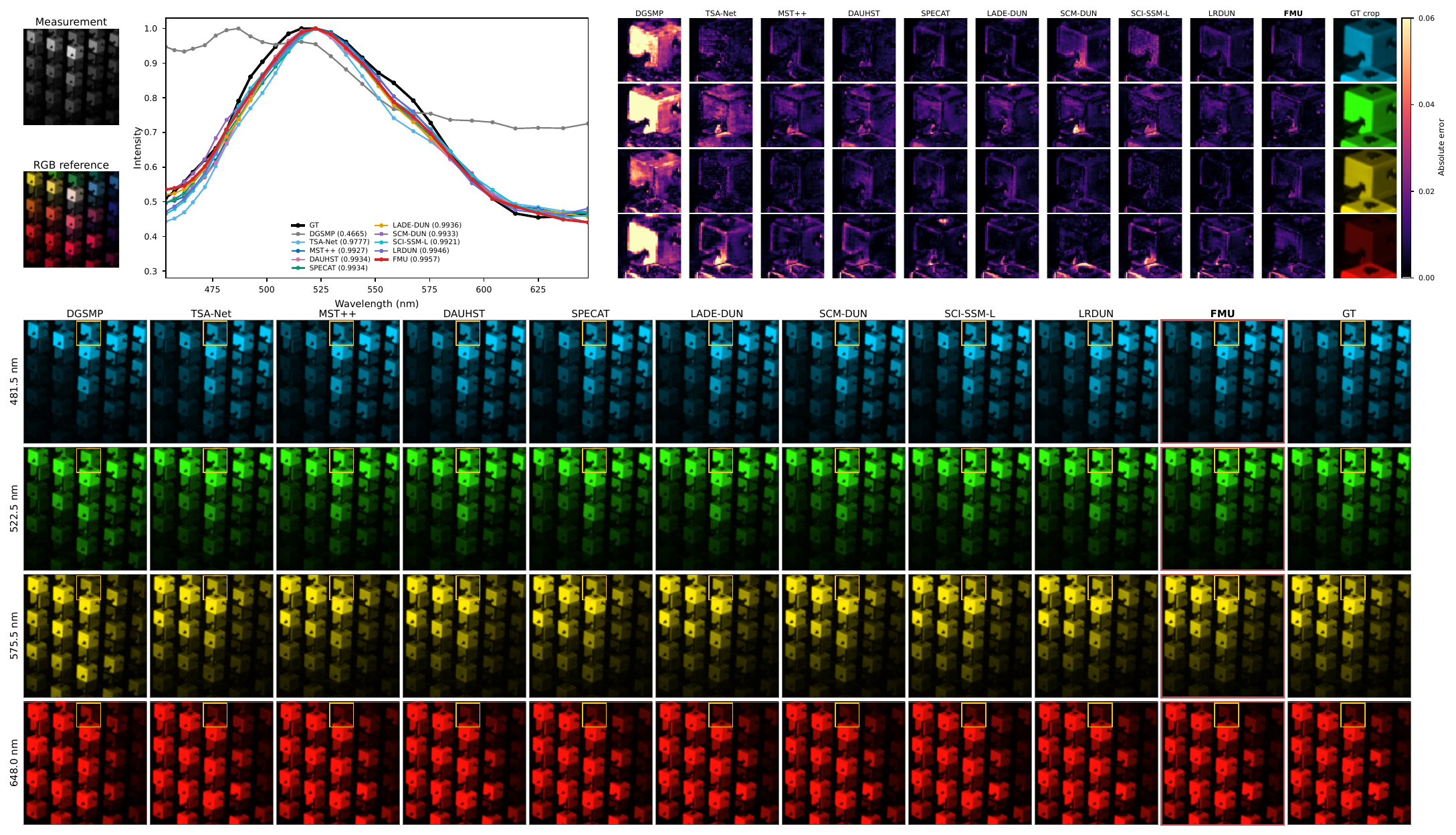}
\caption{Qualitative comparison on the simulated optical-filter-based setting (Scene~2 of the KAIST benchmark). The lower panels show complete reconstructions at 481.5\,nm, 522.5\,nm, 575.5\,nm, and 648.0\,nm with a common display mapping; yellow boxes mark the region of interest (ROI). The upper-right block reports the corresponding absolute-error maps, $|\hat{\bm x}-\bm x|$, followed by the ground-truth crop. All error maps use the same range $[0,0.06]$, where darker values indicate smaller errors. The upper-left panels show the raw measurement, RGB reference, and ROI-mean spectral curves. Each curve is computed after clipping the reconstructed cube to $[0,0.7]$ and independently normalizing its ROI mean to unit peak; the legend reports its correlation with the ground truth. FMU achieves the highest spectral-curve correlation ($0.9957$) among the displayed reconstructions.}
\label{fig:simu}
\end{figure*}

\subsection{Implementation Details}
\label{sec:implementation-details}
We use 10 unfolding stages; the eight intermediate stages share parameters, while the head and tail stages are distinct. We set $\lambda_{mean}=5$. Training uses a batch size of 1, $256\times256$ crops, and clips the gradient norm to 0.2. The flow starts from Gaussian noise with scale $1.0$. RK45~\cite{dormand1980family} integration uses absolute and relative tolerances of $10^{-5}$.

\subsection{Optical-Filter Reconstruction}
\subsubsection{Comparison with State-of-the-Art Methods}
We benchmark FMU against representative reconstruction methods spanning four paradigms: (i) convolutional end-to-end networks---$\lambda$-Net~\cite{miao2019net}, TSA-Net~\cite{meng2020end}, DGSMP~\cite{huang2021deep}, HDNet~\cite{hu2022hdnet}, and BiSRNet~\cite{cai2023binarized}; (ii) recurrent and state-space models---BIRNAT~\cite{cheng2022recurrent} and SCI-SSM-L~\cite{shi2026scissm}; (iii) Transformer-based networks---MST++~\cite{cai2022mstpp}, CST-L-Plus~\cite{cai2022coarse}, and SPECAT~\cite{yao2024specat}; and (iv) deep unfolding networks---ADMM-Net~\cite{ma2019deep}, GAP-Net~\cite{meng2023deep}, DAUHST~\cite{cai2022degradation}, LADE-DUN~\cite{wu2024latent}, SCM-DUN-9stg~\cite{feng2025scmdun}, and LRDUN-9stg~\cite{huang2026lrdun}. Every reconstruction is evaluated on the same ten test cubes using an identical Fabry--P\'{e}rot mask, measurement formation, and metric implementation. All retrained methods use the same 205 training cubes and retain their architecture-specific objectives and optimizers. Table~\ref{table:hsi_scene10_avg} reports per-scene PSNR/SSIM, parameter counts, and venues.

\begin{figure*}[!t]
\centering
\includegraphics[width=\linewidth]{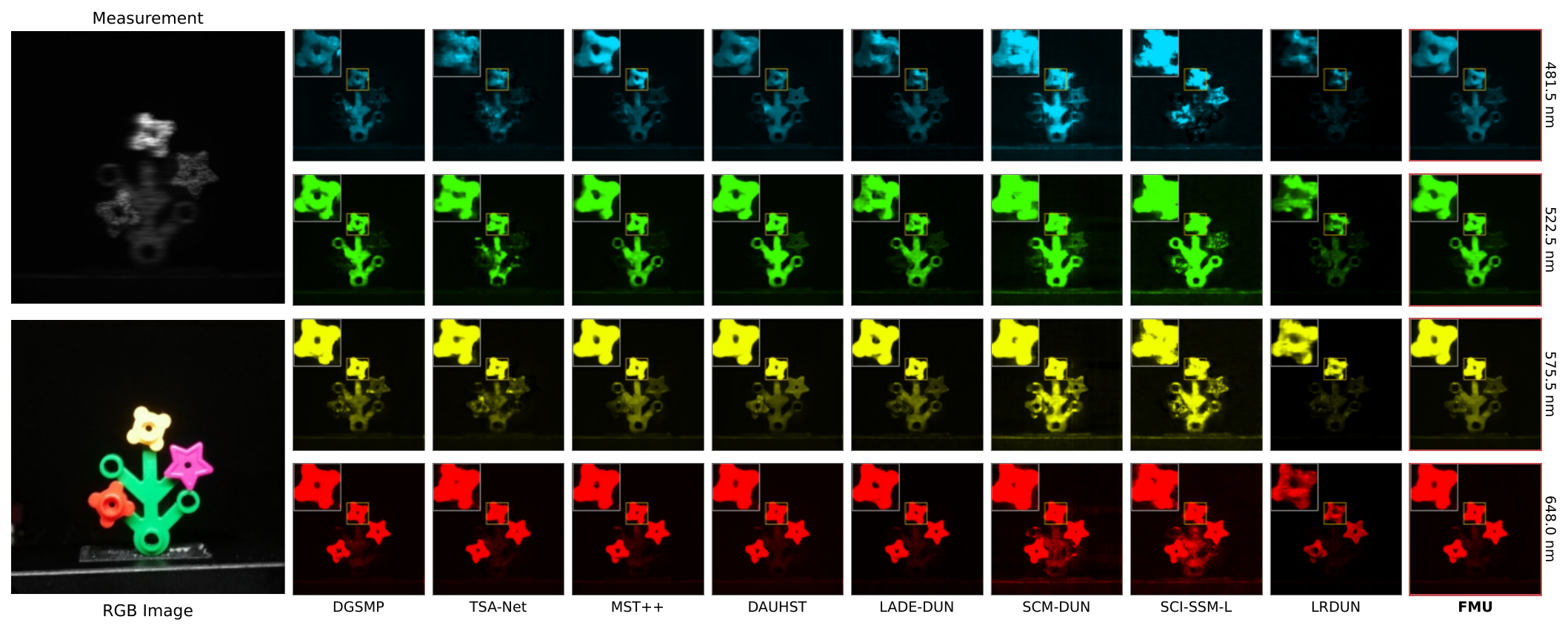}
\caption{Qualitative reconstructions on a real CASSI capture (Scene~1). Each row visualizes one reconstructed spectral channel at 481.5\,nm, 522.5\,nm, 575.5\,nm, and 648.0\,nm. No ground-truth cube exists for this capture, so the panels provide visual references rather than a quantitative ranking.}
\label{fig:real}
\end{figure*}

\begin{table}[!t]
\caption{Multi-metric comparison on the optical-filter KAIST 10-scene benchmark. SAM is reported in radians.}
\label{tab:multi_metric_optical}
\centering
\footnotesize
\renewcommand{\arraystretch}{1.06}
\setlength{\tabcolsep}{2pt}
\begin{tabularx}{\columnwidth}{>{\raggedright\arraybackslash}Xccccc}
\toprule
Method & PSNR$\uparrow$ & SSIM$\uparrow$ & SAM$\downarrow$ & ERGAS$\downarrow$ & MANIQA$\uparrow$ \\
\midrule
BIRNAT & 38.76 & 0.9792 & 0.1064 & 16.79 & 0.6119 \\
BiSRNet & 35.84 & 0.9524 & 0.2126 & 23.12 & 0.5714 \\
CST-L-Plus & 37.73 & 0.9762 & 0.1380 & 19.21 & 0.6180 \\
DAUHST-9stg & 38.81 & 0.9815 & 0.1120 & 17.05 & 0.6273 \\
DGSMP & 32.99 & 0.9471 & 0.1688 & 33.57 & 0.5664 \\
MST++ & 39.21 & 0.9825 & 0.0989 & 16.25 & 0.6256 \\
TSA-Net & 37.68 & 0.9759 & 0.1103 & 18.82 & 0.6124 \\
SPECAT~\cite{yao2024specat} & 40.40 & 0.9859 & 0.0877 & 14.04 & 0.6292 \\
LADE-DUN-10stg~\cite{wu2024latent} & \underline{40.97} & \underline{0.9882} & \underline{0.0708} & \underline{12.70} & \textbf{0.6294} \\
SCM-DUN-9stg~\cite{feng2025scmdun} & 40.49 & 0.9865 & 0.0768 & 13.85 & 0.6238 \\
SCI-SSM-L~\cite{shi2026scissm} & 39.01 & 0.9817 & 0.1040 & 16.56 & 0.6242 \\
LRDUN-9stg~\cite{huang2026lrdun} & 40.85 & 0.9876 & 0.0753 & 13.22 & 0.6174 \\
\textbf{FMU (Ours)} & \textbf{42.13} & \textbf{0.9900} & \textbf{0.0651} & \textbf{11.26} & \underline{0.6293} \\
\bottomrule
\end{tabularx}
\end{table}

\begin{table}[!t]
\caption{Held-out KAIST and ICVL results without fine-tuning.}
\label{tab:kaist_icvl}
\centering
\footnotesize
\renewcommand{\arraystretch}{1.06}
\setlength{\tabcolsep}{3pt}
\begin{tabularx}{\columnwidth}{>{\raggedright\arraybackslash}Xcccc}
\toprule
\rowcolor{color3}
Method & PSNR$\uparrow$ & SSIM$\uparrow$ & SAM$\downarrow$ & ERGAS$\downarrow$ \\
\midrule
\multicolumn{5}{c}{\textit{Held-out KAIST scenes}} \\
\cmidrule(lr){1-5}
LADE-DUN~\cite{wu2024latent} & \underline{43.04} & \underline{0.9939} & \textbf{0.0638} & \underline{14.09} \\
SCM-DUN-9stg~\cite{feng2025scmdun} & 41.88 & 0.9925 & 0.0723 & 16.29 \\
SCI-SSM-L~\cite{shi2026scissm} & 40.28 & 0.9894 & 0.1123 & 19.44 \\
LRDUN-9stg~\cite{huang2026lrdun} & 42.48 & 0.9931 & 0.0693 & 14.99 \\
\rowcolor{light-green}
FMU (Ours) & \textbf{43.44} & \textbf{0.9942} & \underline{0.0645} & \textbf{13.52} \\
\midrule
\multicolumn{5}{c}{\textit{ICVL scenes}} \\
\cmidrule(lr){1-5}
LADE-DUN~\cite{wu2024latent} & \underline{31.92} & \underline{0.9396} & \underline{0.1264} & \underline{81.80} \\
SCM-DUN-9stg~\cite{feng2025scmdun} & 31.00 & 0.9224 & 0.1491 & 107.28 \\
SCI-SSM-L~\cite{shi2026scissm} & 30.15 & 0.9197 & 0.1592 & 109.13 \\
LRDUN-9stg~\cite{huang2026lrdun} & 31.56 & 0.9307 & 0.1420 & 102.26 \\
\rowcolor{light-green}
FMU (Ours) & \textbf{33.58} & \textbf{0.9492} & \textbf{0.1046} & \textbf{73.20} \\
\bottomrule
\end{tabularx}
\end{table}

FMU attains 42.13\,dB PSNR and 0.9900 SSIM on average, the highest values among the compared methods. It exceeds the strongest listed baseline by 1.16\,dB PSNR and 0.0018 SSIM. The 2.78\,M-parameter FMU-S variant reaches 41.94\,dB / 0.9894. In Fig.~\ref{fig:simu}, FMU yields the lowest ROI-averaged absolute error at 522.5\,nm among the displayed methods, and its ROI spectral-curve correlation is $0.9957$.

The per-scene results show that the average gain is distributed across the test set. FMU exceeds LADE-DUN in PSNR on nine of the ten scenes, with gains of 1.74, 1.78, and 1.71\,dB on Scenes~1, 3, and~7, respectively. On Scene~4, LADE-DUN has higher PSNR (47.55 versus 47.14\,dB), while FMU has higher SSIM (0.9955 versus 0.9949). FMU attains the highest SSIM on every scene. At LADE-DUN's parameter count, FMU-S improves the average PSNR by 0.97\,dB.

The full-band images in Fig.~\ref{fig:simu} provide scene context, while the enlarged error maps localize discrepancies within the same ROI. Their shared color scale preserves error-magnitude differences across methods and wavelengths. The peak-normalized curves summarize ROI-mean spectral shape, while the error maps retain absolute intensity discrepancies.

\subsubsection{Spectral and Perceptual Fidelity}
\label{subsec:metrics_fairness}
PSNR and SSIM are complemented with spectral angle mapper (SAM), ERGAS, and MANIQA. SAM measures the angular discrepancy between reconstructed and reference spectra, whereas ERGAS summarizes the band-wise relative reconstruction error. MANIQA is a no-reference perceptual score computed on pseudo-RGB projections; it is therefore reported as an image-quality diagnostic rather than a measure of spectral fidelity. We form each pseudo-RGB image from the 648.0, 551.5, and 462.0\,nm bands in R/G/B order, apply per-image min--max normalization, and use identical preprocessing for every method.

\begin{table}[!t]
\centering
\footnotesize
\renewcommand{\arraystretch}{1.06}
\caption{Simulated-CASSI reconstruction results.}
\label{tab:cassi_results}
\setlength{\tabcolsep}{2pt}
\begin{tabularx}{0.96\columnwidth}{l>{\raggedright\arraybackslash}Xcc}
\toprule
\rowcolor{color3}
Method & Venue & PSNR (dB) & SSIM \\
\midrule
TwIST~\cite{bioucas2007new} & IEEE TIP'07 & 23.12 & 0.669 \\
DNU~\cite{wang2020dnu} & CVPR'20 & 30.74 & 0.863 \\
CASSI-SSL~\cite{he2025selfsupervised} & WACV'25 & 35.52 & 0.944 \\
MST++~\cite{cai2022mstpp} & CVPRW'22 & 35.99 & 0.951 \\
LRSDN~\cite{chen2024hyperspectral} & IEEE TIP'24 & 36.08 & 0.938 \\
RDLUF-MixS2-3stg~\cite{dong2023residual} & CVPR'23 & 37.56 & 0.963 \\
BIRNAT~\cite{cheng2022recurrent} & IEEE TPAMI'23 & 37.58 & 0.960 \\
PADUT-5stg~\cite{li2023pixel} & ICCV'23 & 37.84 & 0.967 \\
PSR-SCI-D~\cite{zeng2025psrsci} & ICLR'25 & 38.14 & 0.967 \\
LADE-DUN-3stg~\cite{wu2024latent} & ECCV'24 & 38.31 & 0.972 \\
DAUHST-9stg~\cite{cai2022degradation} & NeurIPS'22 & 38.36 & 0.967 \\
PADUT-12stg~\cite{li2023pixel} & ICCV'23 & 38.89 & 0.974 \\
DADF-Plus-3~\cite{xu2023degradation} & IEEE TMM'24 & 39.28 & 0.972 \\
RDLUF-MixS2-9stg~\cite{dong2023residual} & CVPR'23 & 39.57 & 0.974 \\
LADE-DUN-10stg~\cite{wu2024latent} & ECCV'24 & 40.09 & 0.979 \\
LRDUN-9stg~\cite{huang2026lrdun} & CVPR'26 & 40.96 & 0.981 \\
\midrule
\rowcolor{light-green}
\textbf{FMU (Ours)} & -- & \textbf{41.02} & \textbf{0.988} \\
\bottomrule
\end{tabularx}
\end{table}

Table~\ref{tab:multi_metric_optical} shows that FMU obtains the highest PSNR and SSIM and the lowest SAM and ERGAS. Its MANIQA score is 0.0001 below the highest listed value, indicating comparable pseudo-RGB perceptual quality.

Relative to LADE-DUN, SAM decreases from 0.0708 to 0.0651 radians and ERGAS from 12.70 to 11.26, corresponding to reductions of 8.1\% and 11.3\%, respectively. These two measures assess different aspects of the reconstructed cube: spectral direction at each pixel and relative error across bands. Their agreement with PSNR supports improved spectral recovery beyond the appearance of the pseudo-RGB projection.

The distinction between spectral and perceptual scores is also visible in the baseline ordering. LRDUN-9stg has lower SAM and ERGAS than SCI-SSM-L (0.0753 versus 0.1040, and 13.22 versus 16.56), although its MANIQA is lower. The pseudo-RGB projection uses three selected bands, whereas SAM and ERGAS assess the full spectral cube. The combined evaluation therefore considers errors in the full cube as well as perceptual structure in its RGB projection.

\subsubsection{Generalization to Held-out KAIST and ICVL Scenes}
To assess generalization beyond the training scenes, we synthesize measurements from additional held-out KAIST images and from ICVL images using the same optical-filter forward model. The ICVL data also introduce different scene and spectral statistics. Neither evaluation split uses fine-tuning.

On the held-out KAIST scenes, FMU obtains the best PSNR, SSIM, and ERGAS and the second-lowest SAM, with LADE-DUN giving the lowest SAM. On ICVL, FMU ranks first under all four metrics, followed by LADE-DUN. These results support cross-dataset generalization under the fixed optical-filter operator used in this experiment.

The PSNR advantage over LADE-DUN is 0.40\,dB on held-out KAIST scenes and 1.66\,dB on ICVL. On ICVL, the gain is accompanied by a decrease in SAM from 0.1264 to 0.1046 and in ERGAS from 81.80 to 73.20. Both reconstruction error and spectral agreement therefore improve without adapting the model to the evaluation scenes.

\subsection{CASSI Reconstruction}
\label{sec:cassi-results}

\subsubsection{Simulated CASSI Benchmark}
Table~\ref{tab:cassi_results} reports published baseline results under the standard CAVE-training/KAIST-testing protocol. FMU is evaluated with the dispersive CASSI operator using a two-pixel wavelength-dependent shift. The $256\times256$ test cubes yield $256\times310$ measurements.

FMU reaches 41.02\,dB PSNR and 0.988 SSIM on this benchmark. LRDUN-9stg obtains the nearest PSNR at 40.96\,dB, while LADE-DUN-10stg reaches 40.09\,dB. Their corresponding SSIM values are 0.981 and 0.979, respectively. The improvement over LADE-DUN is 0.93\,dB in PSNR. These comparisons evaluate reconstruction under wavelength-dependent spatial displacement, complementing the optical-filter results obtained on the same underlying scene splits.

\subsubsection{Real CASSI Measurements}
We reconstruct scenes from the TSA-Net~\cite{meng2020end} capture without fine-tuning on real measurements. Figure~\ref{fig:real} presents reconstructed spectral channels from one representative capture. Because no reference hyperspectral cube is available, these results are used for qualitative inspection rather than quantitative assessment of spatial or spectral fidelity.

In the enlarged ROI, FMU shows a distinct central opening in the upper flower-shaped object at 522.5 and 575.5\,nm, while this opening is less distinct in SCM-DUN and SCI-SSM-L. The full-scene views also reveal wavelength-dependent contrast between the supporting structure and the colored objects.

\subsection{Ablation Study}
Table~\ref{tab:ablation} evaluates the effects of the prior type, mean-velocity weight, and velocity predictor architecture. Unless noted otherwise, the variants use the same 10-stage backbone, training schedule, and evaluation protocol.

\begin{table}[!t]
\centering
\footnotesize
\renewcommand{\arraystretch}{1.06}
\setlength{\tabcolsep}{3pt}
\caption{Ablation studies on the optical-filter KAIST benchmark.}
\label{tab:ablation}
\textbf{(a) Prior type}\par\vspace{2pt}
\begin{tabularx}{0.96\columnwidth}{>{\raggedright\arraybackslash}Xccc}
\toprule
\rowcolor{color3}
Method  & PSNR (dB) & SSIM & FLOPs (G) \\
\midrule
Baseline (no prior) & 40.58 & 0.9878 & \textbf{96.40} \\
\makecell[l]{Latent diffusion\\(LADE-DUN, 2.78M)} & 40.97 & 0.9882 & 96.69 \\
\rowcolor{light-green}
\makecell[l]{Flow matching\\(FMU-S, 2.78M)} & \textbf{41.94} & \textbf{0.9894} & 99.87 \\
\bottomrule
\end{tabularx}
\par\vspace{5pt}
\textbf{(b) Mean-velocity weight}\par\vspace{2pt}
\begin{tabularx}{0.96\columnwidth}{>{\raggedright\arraybackslash}Xcccccc}
\toprule
\rowcolor{color3}
\(\lambda_{mean}\) & 0 & 0.1 & 1 & 5 & 10 & 100 \\
\midrule
PSNR (dB) & 41.90 & 41.97 & 41.99 & \textbf{42.13} & 42.02 & 41.84 \\
SSIM & 0.9887 & 0.9898 & 0.9898 & \textbf{0.9900} & 0.9900 & 0.9897 \\
\bottomrule
\end{tabularx}
\par\vspace{5pt}
\textbf{(c) Velocity predictor architecture}\par\vspace{2pt}
\begin{tabularx}{0.96\columnwidth}{>{\raggedright\arraybackslash}Xccc}
\toprule
\rowcolor{color3}
Variant & PSNR (dB) & Params (M) & FLOPs (G) \\
\midrule
MLP & 41.95 & \textbf{2.78} & 99.87 \\
gMLP & 42.05 & 4.36 & 103.27 \\
Tiny Transformer & 41.90 & 9.09 & 109.45 \\
\rowcolor{light-green}
SimpleCNN (Ours) & \textbf{42.13} & 4.09 & \textbf{98.84} \\
\bottomrule
\end{tabularx}
\end{table}

\textbf{Effect of the generative prior.}
Table~\ref{tab:ablation}(a) compares prior formulations with the unfolding architecture and training protocol held fixed. LADE-DUN and FMU-S share the same 10-stage topology, denoiser, training data and schedule, and 2.78\,M parameter budget; the generative model and its prior-learning objective are changed. FMU-S reaches 41.94\,dB / 0.9894, compared with 40.97\,dB / 0.9882 for latent diffusion and 40.58\,dB / 0.9878 without a prior. The 0.97\,dB gain over diffusion is obtained without increasing parameter count or changing the reconstruction backbone.

\textbf{Weight of the mean-velocity constraint.}
Table~\ref{tab:ablation}(b) sweeps $\lambda_{mean}\in\{0, 0.1, 1, 5, 10, 100\}$. Removing the constraint drops PSNR by 0.23\,dB relative to the best setting, $\lambda_{mean}=5$, while the largest weight reduces PSNR by 0.29\,dB. Both weights 5 and 10 achieve 0.9900 SSIM, but the former gives higher PSNR. These results favor a moderate weight on the mean error across channels. The aggregate term complements samplewise endpoint matching, but placing excessive weight on the channel-wise average can reduce the final reconstruction accuracy.

\textbf{Velocity predictor architecture.}
Table~\ref{tab:ablation}(c) compares four velocity-predictor architectures with the reconstruction denoiser held fixed. SimpleCNN obtains the highest PSNR (42.13\,dB) and lowest FLOPs (98.84\,G) among the evaluated configurations. The Tiny Transformer configuration uses 9.09\,M parameters but reaches 41.90\,dB, showing that a larger predictor does not necessarily improve the reconstruction. Compared with gMLP, SimpleCNN gives slightly higher PSNR with fewer parameters and FLOPs. The MLP offers the smallest configuration at 2.78\,M and reaches 41.95\,dB. We therefore use SimpleCNN for the full model, while the MLP provides a lower-parameter alternative for the same latent-prediction task.

\raggedbottom
\section{Conclusion}
We proposed FMU, a deep unfolding framework that combines a measurement-conditioned flow-matching prior with a sensing-model-guided measurement update for snapshot hyperspectral reconstruction. A compact clean-HSI latent target is learned before a conditional velocity field is optimized to generate the prior from the measurement. The generated latent is sampled once and shared across unfolding stages, while the mean-velocity term regularizes the channel-wise aggregate motion of the learned field. FMU obtains 42.13\,dB PSNR and 0.9900 SSIM on the optical-filter benchmark, improving upon LADE-DUN by 1.16\,dB in PSNR under the matched data and sensing protocol. Under the same optical-filter operator, FMU is also evaluated on held-out KAIST and ICVL scenes without fine-tuning. The simulated CASSI experiment provides quantitative evaluation under a dispersive sensing operator, whereas the real-CASSI experiment provides qualitative reconstructions in the absence of reference hyperspectral cubes. Future work will study faster sampling and reconstruction under sensor noise and mask mismatch.

\ifCLASSOPTIONcaptionsoff
  \newpage
\fi

% references section
\clearpage
\bibliographystyle{IEEEtran}
\bibliography{refs}

% biography section
\vspace{-4mm}
\begin{IEEEbiography}[{\bioheadshot{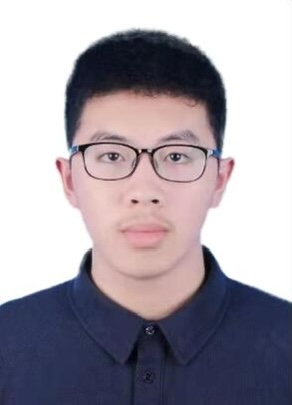}}]{Yi Ai}
received the B.E. degree from Shanghai Jiao Tong University, Shanghai, China. He is currently pursuing the Ph.D. degree with the School of Computer Science, Shanghai Jiao Tong University. His research interests include computer vision, particularly hyperspectral imaging and low-light image enhancement.
\end{IEEEbiography}

\begin{IEEEbiography}
[{\bioheadshot{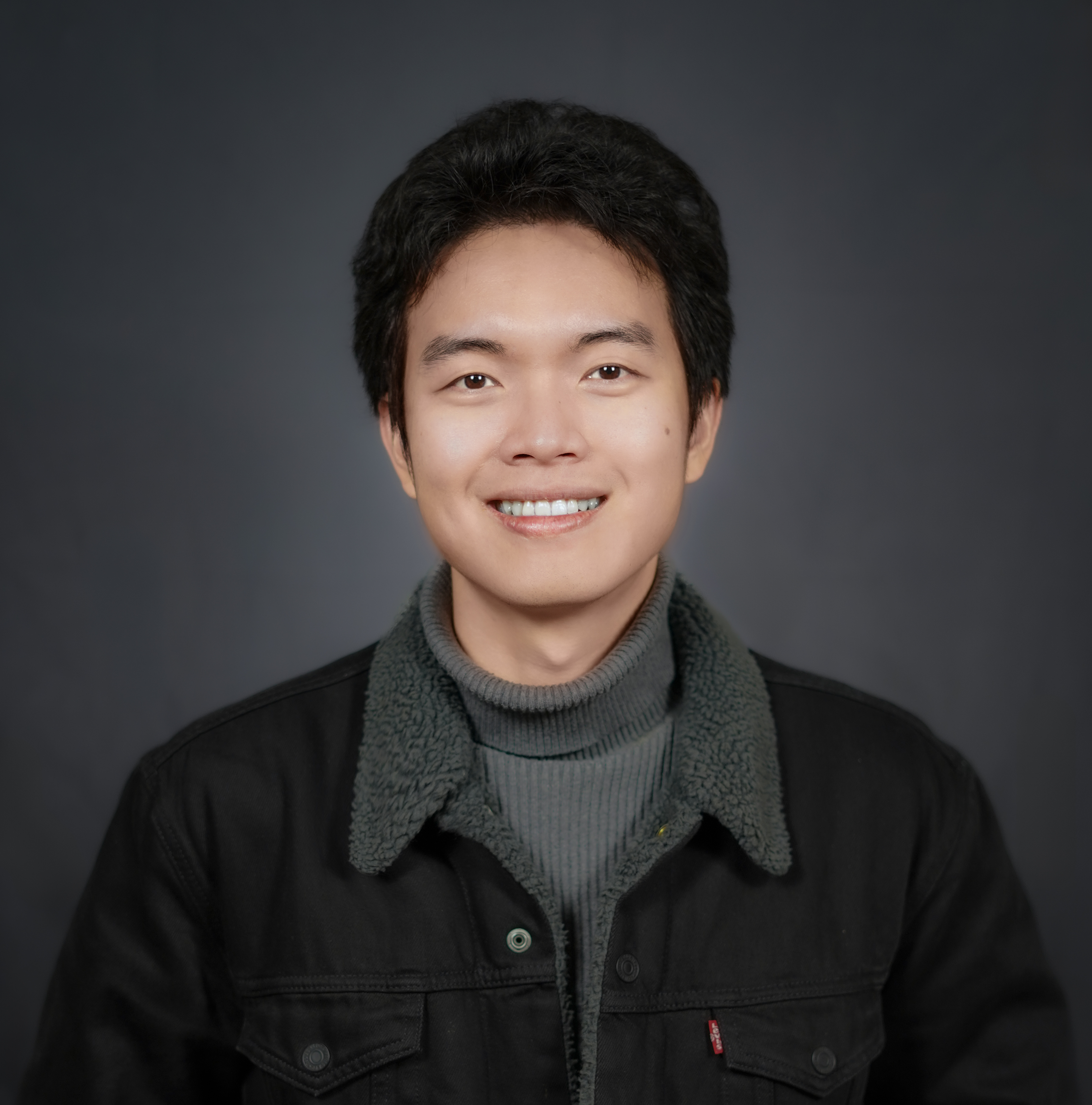}}]{Yuanhao Cai}
is a third-year Ph.D. student at Johns Hopkins University, advised by Professor Alan Yuille. His research focuses on generative AI, including 3D generation, as well as image/video generation and editing. He has published over 20 papers at top-tier conferences, including CVPR, ICCV, ECCV, ICLR, ICML, and NeurIPS. He was also recognized as a Stanford/Elsevier World’s Top 2\% Scientist in 2025.
\end{IEEEbiography}

\begin{IEEEbiography}[{\bioheadshot{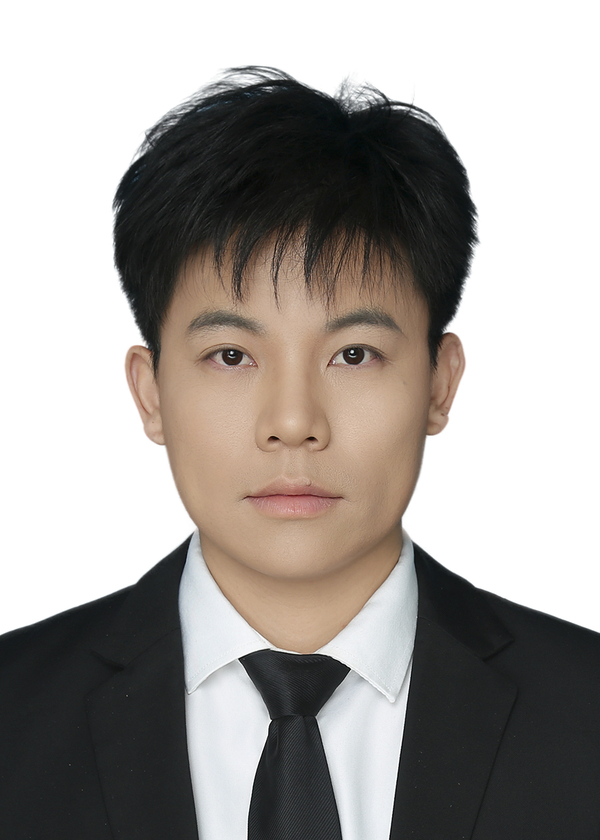}}]{Yulun Zhang}
received a B.E. degree from the School of Electronic Engineering, Xidian University, China, in 2013, an M.E. degree from the Department of Automation, Tsinghua University, China, in 2017, and a Ph.D. degree from the Department of ECE, Northeastern University, USA, in 2021. He is an associate professor at Shanghai Jiao Tong University, Shanghai, China. He was a postdoctoral researcher at Computer Vision Lab, ETH Zürich, Switzerland. His research interests include image/video restoration and synthesis, model compression, multimodal computing, large language model, and computational imaging. He is/was an Area Chair for CVPR, ICCV, ECCV, NeurIPS, ICML, ICLR, IJCAI, ACM MM, and AAAI.
\end{IEEEbiography}

\vspace{-4mm}
\begin{IEEEbiography}[{\bioheadshot{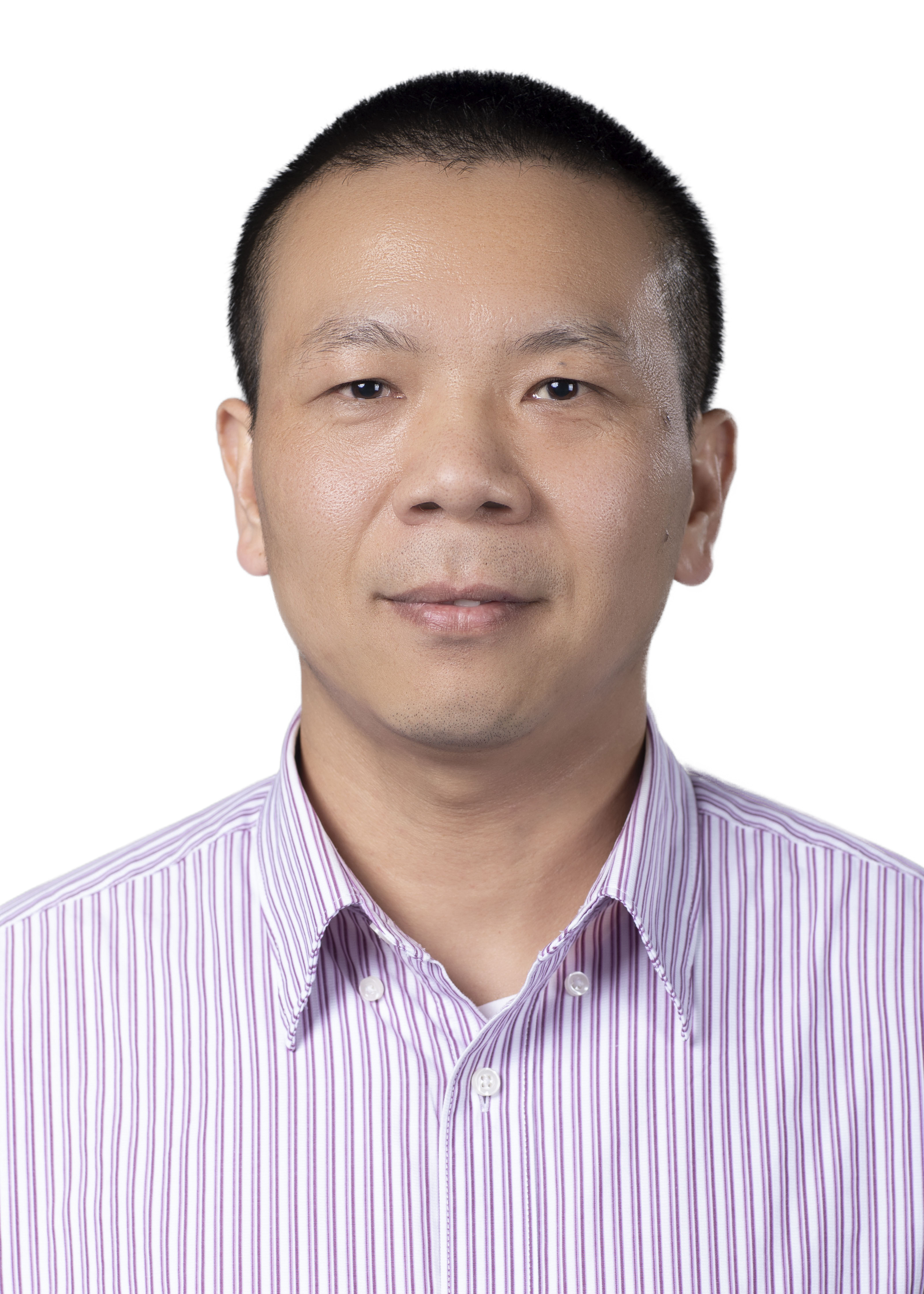}}]{Xin Yuan}
(Senior Member,~IEEE) received the B.Eng. and M.Eng. degrees from Xidian University, in 2007 and 2009, respectively, and the Ph.D. from the Hong Kong Polytechnic University, in 2012. He is currently an Associate Professor at Westlake University. He was a video analysis and coding lead researcher at Bell Labs, Murray Hill, NJ, USA from 2015 to 2021. He was a Postdoctoral Associate at Duke University from 2012 to 2015. His research interests are in signal processing, computational imaging and machine learning. He has been the receiving editor of Optics and Laser Technology since 2024, the Senior Associate Editor of IEEE Transactions on Image Processing since 2026, the Associate Editor of Pattern Recognition since 2019, the Topic Editor of Chinese Optics Letters since 2021, and the Deputy Editor of Advanced Imaging since 2026. He is an Optica Fellow.
\end{IEEEbiography}

\vspace{-4mm}
\begin{IEEEbiography}[{\bioheadshot{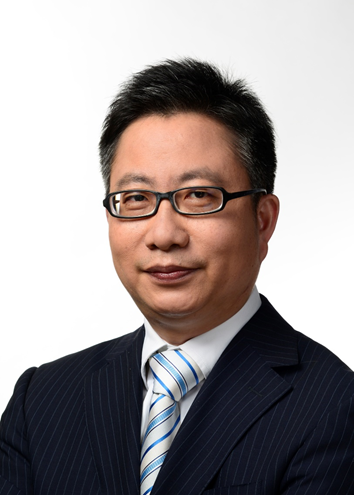}}]{Xiaokang Yang}
(Fellow,~IEEE) received the B.S. degree from Xiamen University, Xiamen, China, in 1994, the M.S. degree from Chinese Academy of Sciences, Shanghai, China, in 1997, and the Ph.D. degree from Shanghai Jiao Tong University, Shanghai, in 2000. From 2000 to 2002, he was a Research Fellow with the Centre for Signal Processing, Nanyang Technological University, Singapore. From 2002 to 2004, he was a Research Scientist with the Institute for Infocomm Research (I2R), Singapore. From 2007 to 2008, he visited the Institute for Computer Science, University of Freiburg, Germany, as an Alexander von Humboldt Research Fellow. He is currently a Distinguished Professor with the School of Electronic Information and Electrical Engineering, Shanghai Jiao Tong University. His current research interests include image processing and communication, computer vision, and machine learning. He is an Associate Editor of IEEE Transactions on Multimedia and a Senior Associate Editor of IEEE Signal Processing Letters.
\end{IEEEbiography}

\end{document}